\documentclass{article}

\usepackage[final]{corl_2020} % Uncomment for the camera-ready ``final'' version.
\usepackage{graphicx}
\usepackage{booktabs}       % professional-quality tables
\usepackage{amsfonts}       % blackboard math symbols
\usepackage{amsmath}
\usepackage{amssymb}
\usepackage{nicefrac}       % compact symbols for 1/2, etc.
\usepackage{microtype}      % microtypography
\usepackage{float}
\usepackage{multirow}

\usepackage{graphicx}
\usepackage{subfig}
\usepackage[capitalise]{cleveref}
\usepackage[font=footnotesize,labelfont=bf]{caption}

\usepackage{amsmath,amsfonts,bm}

\def\Figref#1{Fig.~\ref{#1}}

\def\Tabref#1{Table~\ref{#1}}
\def\Appref#1{Appendix~\ref{#1}}
\def\eqref#1{equation~\ref{#1}}
\def\1{\bm{1}}

\DeclareMathAlphabet{\mathsfit}{\encodingdefault}{\sfdefault}{m}{sl}
\SetMathAlphabet{\mathsfit}{bold}{\encodingdefault}{\sfdefault}{bx}{n}

\title{Visual Imitation Made Easy}
\author{Sarah Young\textsuperscript{1}  \qquad Dhiraj Gandhi \textsuperscript{2} \qquad Shubham Tulsiani\textsuperscript{2} \vspace{0.05in} \\ \vspace{0.1in} \textbf{Abhinav Gupta\textsuperscript{2,3} \qquad Pieter Abbeel\textsuperscript{1} \qquad Lerrel Pinto\textsuperscript{1,4}} \\ 
 \textsuperscript{1}University of California, Berkeley  \qquad \textsuperscript{2}Facebook AI Research  \\ \vspace{0.1in} \textsuperscript{3}Carnegie Mellon University \qquad  \textsuperscript{4}New York University\\ 
{\tt \small \href{https://dhiraj100892.github.io/Visual-Imitation-Made-Easy/}{Project Page}}
}

\begin{document}
\maketitle

%===============================================================================

\begin{abstract}
Visual imitation learning provides a framework for learning complex manipulation behaviors by leveraging human demonstrations. However, current interfaces for imitation such as kinesthetic teaching or teleoperation prohibitively restrict our ability to efficiently collect large-scale data in the wild. Obtaining such diverse demonstration data is paramount for the generalization of learned skills to novel scenarios. In this work, we present an alternate interface for imitation that simplifies the data collection process while allowing for easy transfer to robots. We use commercially available reacher-grabber assistive tools both as a data collection device and as the robot's end-effector. To extract action information from these visual demonstrations, we use off-the-shelf Structure from Motion (SfM) techniques in addition to training a finger detection network. We experimentally evaluate on two challenging tasks: non-prehensile pushing and prehensile stacking, with 1000 diverse demonstrations for each task. For both tasks, we use standard behavior cloning to learn executable policies from the previously collected offline demonstrations. To improve learning performance, we employ a variety of data augmentations and provide an extensive analysis of its effects. Finally, we demonstrate the utility of our interface by evaluating on real robotic scenarios with previously unseen objects and achieve a 87\% success rate on pushing and a 62\% success rate on stacking. Robot videos are available at our \href{https://dhiraj100892.github.io/Visual-Imitation-Made-Easy/}{project website}.

\end{abstract}

\keywords{Imitation learning, Generalization, Visual observations.} 

%===============================================================================

% \input{math_commands}
\section{Introduction}
A powerful technique to learn complex robotic skills is to imitate them from humans~\cite{piaget2013play,meltzoff1977imitation,meltzoff1983newborn, tomasello1993imitative}. Recently, there has been a growing interest in learning such skills from visual demonstrations, since it allows for generalization to novel scenarios~\cite{zhang2018deep, stadie2017third}. Prominent works in Visual Imitation Learning~(VIL) have demonstrated utility in intricate manipulation skills such as pushing, grasping, and stacking~\cite{zhang2018deep,DBLP:journals/corr/abs-1802-09564}. However, a key bottleneck in current imitation learning techniques is the use of interfaces such as kinesthetic teaching or teleoperation, which makes it harder to collect large-scale manipulation data. But more importantly, the use of such interfaces leads to datasets that are constrained to be in restrictive lab settings. Resulting in-lab demonstrations often contain little to no variability in objects or environments, which severely limits the generalizability of the learned skills in novel, previously unseen situations~\cite{gupta2018robot}.

\begin{figure*}[t!]
	\centering
	\includegraphics[width=\textwidth]{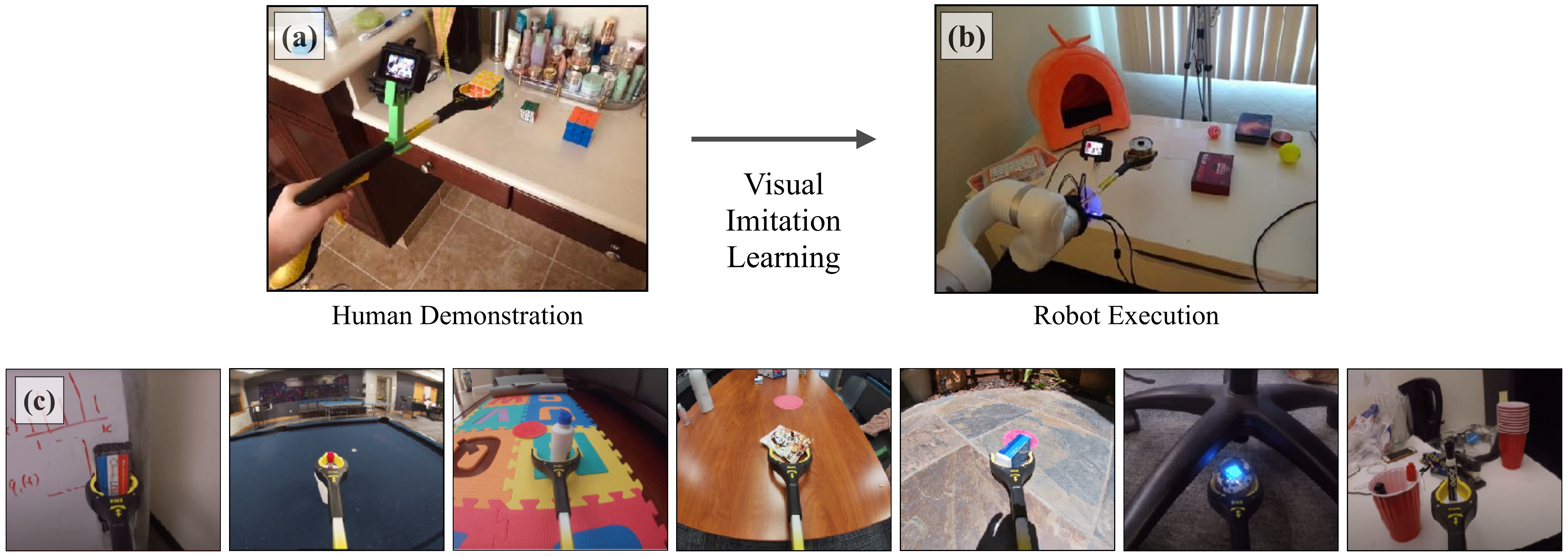}
	\caption{In this work we present a framework for visual imitation learning, where demonstrations are collected using commercially available reacher-grabber tools~(a). This tool is also instrumented as an end-effector and attached to the robot~(b). This setup allows us to collect and learn from demonstration data across diverse environments~(c), while allowing for easy transfer to our robot.} 
	\label{fig:teaser}
\end{figure*}

% How can one simplify data collection for imitation learning to allow both data collection at scale and real world diversity? 
It is thus important to find a way to simplify data collection for imitation learning to allow both data collection at scale and real world diversity. What we need is a cheap interface (for prevalence), which can be intuitively controlled (for efficiency). Interestingly, one of the cheapest `robots' that is highly prevalent, easy to control, and requires little to no human training is the reacher-grabber depicted in \Figref{fig:teaser}. This assistive tool is commonly used for grasping trash among other activities of daily living and has recently been shown to be a scalable interface for collecting grasping data in the wild by~\citet{song2020grasping}. However, unlike teleoperation~\cite{zhang2018deep} or kinesthetic~\cite{sharma2018multiple} interfaces where the demonstrations are collected on the same platform as the robot, assistive tools are significantly different from robotic manipulators. \citet{song2020grasping} bridges this gap by first extracting grasp points from demonstrations and then transferring them to robot in order to achieve closed-loop grasping of novel objects. A key problem, however, still lies in scaling this to enable imitation of general robotics tasks. One possible solution is to extract full tool configuration and learn a mapping between grabber and the robot hardware. An alternative is to run domain adaptation based techniques for transfer. However, effectively using such techniques in robotics is still an active area of research~\cite{stadie2017third,sermanet2016unsupervised}. Instead, why not simply use the assistive tool as an end-effector?

In this work, we propose an alternate paradigm for providing and learning from demonstrations. As seen in \Figref{fig:teaser}~(a,c), the user collects Demonstrations using Assistive Tools (DemoAT) to solve a task. During the collection of this data, visual RGB observations are collected from a camera mounted on the DemoAT tool. Given these visual demonstrations, we extract tool trajectories using off-the-shelf Structure from Motion~(SfM) methods and the gripper configuration using a trained finger detector. Once we have extracted tool trajectories, corresponding skills can be learned using standard imitation learning techniques. Particularly, we employ simple off-the-shelf behavior cloning. Finally, these skills can be transferred to a robot that has the same tool setup as the end-effector. Having the same end-effector as the demonstration tool coupled with a 6D robotic control (\Figref{fig:teaser}~(b)) allows for a direct transfer of learning from human demonstrations to the robot.

To study the effectiveness of this tool, we focus on two challenging tasks: (a) non-prehensile pushing~\cite{lynch1996nonprehensile, ruggiero2018nonprehensile}, and (b) prehensile stacking~\cite{zhang2009autonomous,furrer2017autonomous}. For both tasks, we collect 1000 demonstrations in multiple home and office environments with various different objects. This diversity of data in objects and environments allows our learned policies to generalize and be effective across novel objects. Empirically, we demonstrate a baseline performance of 62.5\% in pushing and 29.2\% in stacking with naive behavioral cloning on our robot with objects previously unseen in the demonstrations. We employ random data augmentations such as random crops, jitter, cuts, and rotations to significantly improve pushing performance to 87.5\% and stacking performance to 62.5\%. Finally, we analyze the effects of diversity to demonstrate the need for large-scale demonstration data in the wild. 

In summary, we present three key contributions in this work. First, we propose a new interface for visual imitation learning that uses assistive tools to gather diverse data for robotic manipulation, including an approach for collecting grabber 3-D trajectories and gripper transitions.  Second, we demonstrate the utility of this framework on pushing and stacking previously unseen objects, with a success rate of 87.5\% and 62.5\% respectively. Finally, we present a detailed study on the effects of data augmentations in learning robotic skills, and demonstrate how the combination of random `crops', `rotations' and `jitters' significantly improve our policies over other augmentations. %Our platform design, software, and data will be \href{https://sites.google.com/view/visual-imitation-made-easy}{publicly released} and is attached in the supplementary material.

\section{Related Work}
In this section, we briefly discuss prior research in the context of our work. For a more comprehensive review of imitation learning, we point the readers to \citet{argall2009survey}.

\textbf{Interfaces for Imitation:}
In imitation learning, a robot tries to learn skills from demonstrations provided by the expert. There are various interfaces through which these demonstration can be recorded. One option is teleoperation, in which the human controls the robot using a control interface. This method has been successfully applied to a large range of robotic tasks including flying a robotic helicopter \cite{AndreNgHelicopter}, grasping objects \cite{pook1993recognizing, sweeney2007model}, navigating robots through cluttered environments \cite{inoue1999acquisition, smart2002making, ross2013learning}, and even driving cars \cite{bojarski2016end}. Teleoperation has been successful in solving a wide variety of tasks because of the availability of control interfaces through which human operators can perform high-quality maneuvers. However, it is challenging to devise such interfaces for robotic manipulation \cite{handa2019dexpilot}. Kinesthetic demonstrations, in which the expert actively controls the robot arm by exerting force on it, is an effective method \cite{akgun2012keyframe, sharma2018multiple} of collecting robot manipulation demonstrations for playing ping pong \cite{muelling2014learning} and cutting vegetables \cite{lenz2015deepmpc}. However, for visuomotor policies, which map from pixels to actions, these demonstrations are inappropriate due to the undesirable appearance of human arms. One way to overcome this problem is by mounting an assistive tool on the robot end effector that is being used to record demonstration in isolation \cite{song2020grasping}. We take this idea a step further by using it as an end-effector on the robot as well. This eliminates the domain gap between the human-collected demonstrations and the robot executions, which enables easier imitation.

%One way to overcome this problem is by mounting an assistive tool on the robot end effector that is being used to record demonstration in isolation \cite{song2020grasping}

% notes on song paper: It has a servo motor, compute stick, and one RGBD camera.  
% -  we use a similar approach of recovering true actions. 
% Another difference is that their goal focuses on grasping objects rather than place them/do other tasks.
%  uses off-policy Q-learning to train visual grasping value functions. 

% Similar to ~\citet{song2020grasping}, we use a cost-effective, sturdy, and readily available reacher-grabber to collect imitation data. However, our setup is much simpler and nondestructive. We simply have a 3D printed part mounted on stick to hold a GoPro camera. Instead of collecting RGB-D data \citet{song2020grasping}, we only collect RGB data and then run it through structure from motion to get action information. The cost of our interface can be further reduced by replacing GoPro camera by mobile phone camera.

% \textcolor{red}{Describe different interfaces for doing imitation. Highlight limitations. Introduce work by Song et al.}

\textbf{Behavior Cloning in Imitation:}
Behavior cloning is the simplest form of imitation learning, where the agent learns to map observations to actions through supervised learning. It has been successfully applied in solving a wide range of tasks including playing games~\cite{ross2011reduction}, self-driving~\cite{bojarski2016end}, and navigating drones through cluttered environments~\cite{ross2013learning}. However, it has not been widely applicable to learning visuomotor policies for robotic manipulation tasks due to unwanted visual artifacts collected in kinesthetic demonstrations. To overcome this problem, \citet{zhang2018deep} propose a Virtual Reality (VR) setup to collect robot manipulation data. They showed that behavior cloning can be used to learn complex manipulation tasks, such as grasping and placing various objects. There have also been recent efforts to imitate from visual demonstrations collected from a different space e.g. from a different viewpoint or an agent with a different embodiment~\cite{stadie2017third,sermanet2016unsupervised} from the robot. This is a promising direction as it allows for data collection outside the lab. However, learning from such demonstrations is still an active research problem, as there is a significant domain gap between training and testing. In our setup, we use behavior cloning to learn challenging tasks such as pushing and stacking. But instead of relying on a costly VR setup which can only be deployed in constrained lab environments, we rely on cheap assistive tools to collect diverse data in the wild. Further, we eliminate the domain gap present in previously mentioned lines of work by attaching the same tool on the robot to match the demonstration and imitation space.

\textbf{Data Augmentation in Learning:}
Data augmentation is widely used in machine learning to inject additional knowledge in order to overcome the challenges of overfitting. This technique has been shown to greatly benefit deep learning systems for computer vision. Its use can be found as early as LeNet-5~\cite{lecun1998gradient}, which was used to classify hand written digits. In AlexNet~\cite{krizhevsky2012imagenet}, data augmentations such as random flip and crop were used to improve the classification accuracy. More recently, learning augmentation strategies from data has emerged as a new
paradigm to automate the design of augmentation  \cite{cubuk2019autoaugment, ho2019population, zoph2019learning}. For unsupervised and semi-supervised learning, several unsupervised data augmentation techniques
have been proposed \cite{berthelot2019mixmatch,xie2019unsupervised}. It has also been extensively used in context of RL, where domain randomization is proposed to transfer learning from simulation to real world~\cite{sadeghi2016cad2rl,tobin2017domain, pinto2017asymmetric}. Although the effects of augmentations have been extensively studied in image-based RL~\cite{laskin2020reinforcement,drq}, to the best of our knowledge, we are the first to study the effects of data augmentations in real-robot applications.
% for RL and introduce two
% new data augmentations - random translate and random amplitude scale. In our setup, we followed the data augmentation strategies proposed by RAD to better transfer our policies to robot.

% \textcolor{red}{A lot of recent papers in this area. Good to cite all of them. Highlight simplicity from data augmentations as opposed to algorithmic advances.}
\section{Method}
In this section, we describe the Demonstrations with Assistive Tools (DemoAT) framework for collecting visual demonstrations, along with our pipeline for imitation learning.

\subsection{The DAT imitation framework}
% \textcolor{red}{Details on the setup. Obtaining labels for imitation. Setup of the robot. Highlight no-domain gap in both action space and observation space.}

\begin{figure*}[t!]
	\centering
	\includegraphics[width=\textwidth]{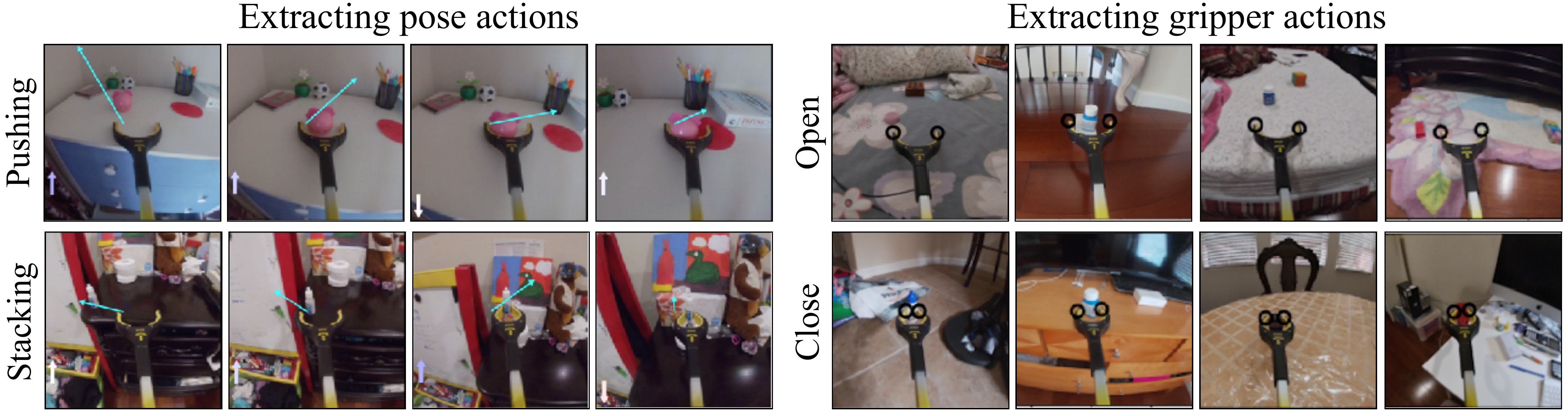}
	\caption{Extracting labels: (a) COLMAP translation arrows are shown for pushing and stacking. The center blue arrow shows movement in the transverse plane of the camera, while the color map arrow in the bottom left corner shows up-down movement. (b) Gripper finger predictions from our finger detection network along with open and close labels for the gripper configuration.} 
	\label{fig:labels}
\end{figure*}

\textbf{Demonstration Tool:} Our DemoAT setup is built around a plastic 19-inch RMS assistive tool~\cite{grabber} and a RGB camera~\cite{gopro} to collect visual data. We attach a 3D printed mount above the stick to hold the camera in place. At the base of the reacher-grabber, there is a lever to control the opening and closing of the gripper fingers. To collect demonstrations, a human user uses the setup shown in \Figref{fig:teaser}~(a), which allows the user to easily push, grab and interact with everyday objects in an intuitive manner. Examples of demonstrations can be seen in \Figref{fig:teaser}~(c) and \Figref{fig:labels}. Since a demonstration collected with DemoAT is visual, it can be represented as a sequence of images $\{I_t\}_{t=0}^{T}$. 

\textbf{Robot End-effector:}
The tool is attached on a 7DoF robot arm with a matching camera and mount setup (\Figref{fig:teaser}~(b)). However, to actuate the fingers, we will need to create an actuating mechanism. Through a compact, lightweight and novel mechanism, we replace the lever from the original reacher grabber tool with a controllable interface. Details on this mechanism are presented in \Appref{app:tool}. While we use an xArm7 robot~\cite{xArm} as our robotic arm, we note that this end-effector setup can be attached to any standard commercial-grade robotic arm.

\subsection{Extracting actions from Visual Demonstrations} 
Although our demonstration tool provides a robust and reliable interface to collect visual demonstrations, our framework in itself does not have explicit sensors to collect information about actions such as the end-effector's motion or the finger locations. For effective imitation learning, this information about the `actions' taken by the human demonstrator is crucial. To address this, we recover 6DoF poses of the tool using Structure-from-Motion (SfM) reconstruction. Specifically, we use the publicly available COLMAP~\cite{colmap1,colmap2} software for SfM. Once we have the end-effector pose $p_t$ for every image $I_t$, we extract the relative translation and rotation $\Delta p_t$ between consecutive frames and use them as the action for training. As SfM only allows us to recover pose up to a scaling factor, we normalize $\Delta p_t$ across the trajectory to account for this ambiguity.
% However, this method of labeling is not perfect. We find that COLMAP is significantly less accurate in lightly textured, clean, and high dynamic range environments. In the tasks we study, objects in the scene are not stationary, another reason for noisy labeled data. Furthermore, our labels have no consistent distance metric in the real world. To combat these problems, we carefully clean the data and normalize translation labels. 

COLMAP gives us the relative change in pose across frames, however, we also need to obtain the finger configurations for tasks that require moving the fingers. To do this, we use a neural network that extracts the location of the gripper fingers in our observations. This network is trained on a small human-labeled dataset of 155 frames from the DemoAT setup. Given these gripper finger locations predicted by our gripping model, we can generate labels for ``close" or ``open" states $g_t \in \{0,1\}$. For this we track the distance between fingers. If distance falls below a threshold, we annotate them as ``close", otherwise ``open". Through this procedure we can now obtain visual demonstrations with actions $a_t$, which is represented as $(o_t, a_t=(\Delta p_t, g_{t+1}))_{t=0}^T$. Note that the grasping action at a given timestep is the grasp state at the next timestep. Visualizations of actions can be seen in \Figref{fig:labels}.

% We explore the effects of noisy labels in \Secref{sec:42}.
% Visualizations of the COLMAP, grasp labels etc.
% Performance effects with poor COLMAP extraction?

\textbf{Accuracy of reconstructed actions:} Our method for extracting labels can be noisy. Specifically, COLMAP reconstructions are significantly less accurate in lightly textured, clean, and high dynamic range scenes. However, since our demonstrations are collected in cluttered real-world scenarios, our reconstructions are reasonably accurate for the purposes of learning. Nevertheless, to reduce the effect of noisy action labels, we visually inspect the reconstructions and discard $\sim 6\%$ of aberrant demonstrations. The model we use to generate grasping actions by detecting finger achieves $\sim 95\%$ accuracy on held-out testing set, which is empirically sufficient for downstream learning.

% \textcolor{red}{COLMAP for 6DOF movement, neural network for fingers, or grasping -- Show figure of results from both. Discuss where does the model fail - clean environments.}

\subsection{Imitation from visual demonstrations}

\textbf{Visual behavior cloning:}
We learn a policy using straightforward behavioral cloning~\cite{pomerleau1989alvinn, ross2011reduction}. With the DAT imitation framework, we collect observation-action pairs $D = \{(o_t, a_t)\}$, where $o_t$ is an image and $a_t$ is the action to get from $o_t$ to $o_{t+1}$. Using supervised learning, our policy learns a function $f(o_t, a_t) $ that maps observations $o_t$ to actions $a_t$. 

The input to the network is a single image $I_t \in \mathbb{R}^{3 x 224 x 224}$. The network outputs actions consisting of (a) a translation vector $ x_t \in \mathbb{R}^3 $ (b) a 6D representation of rotation $w_t \in \mathbb{R}^6$. We train on a 6D rotation representation \cite{6drotation} because it is continuous in the real Euclidean space and thus more suitable for learning as opposed to more commonly used axis-angle and quaternion based representations. 

Our network architecture consists of a CNN with a set of fully connected layers. The convolutional part of the network comprises of the first five layers of the AlexNet followed by an additional convolutional layer. The output from the convolutions are fed into a set of two fully connected layers and projected to a 3D translation vector. To obtain predicted rotations, we concatenate the convolutional representation of the image with the predicted translations and feed this through a fully connected layer before projecting to a 6D rotation vector.
For tasks that require using the gripper, we train an additional classification model that takes in an image $I_t \in \mathbb{R}^{3 x 224 x 224}$ and outputs a gripper open/close label $g_{t+1} \in \{1, 0\} $. Additional details on training are presented in \Appref{app:training}.

% \begin{figure*}[hb]
% 	\centering
% 	\includegraphics[width=\textwidth, height=0.3\textwidth]{images/fig3.pdf}
% 	\caption{network architecture that show network. Integrate both $\Delta p_t$ and $g_t$}. 
% 	\label{fig:arch}
% \end{figure*}

\textbf{Data augmentations for imitations:} 
To improve the performance of our networks with limited data, we experiment with using the following data augmentations in training~\cite{laskin2020reinforcement,drq,simclr}:
% . We consider the following augmentations:

\begin{itemize}

\item Color Jitter: Randomly adds up to $\pm 20\%$ random noise to the brightness, contrast and saturation of each observation.

\item Crop: Randomly extracts a 224 x 224 patch from an original image of size 240 x 240. 

\item Cutout-color[rad]: Randomly inserts a colored box of size [10, 60] into the image. 

\item Rotation: Randomly rotates original image [-5, 5] degrees. 

\item Horizontal Reflection: Mirrors image across the y-axis. Action labels are reflected as well.
\end{itemize}

\section{Experiments}

\begin{figure*}[t!]
	\centering
	\includegraphics[width=\textwidth]{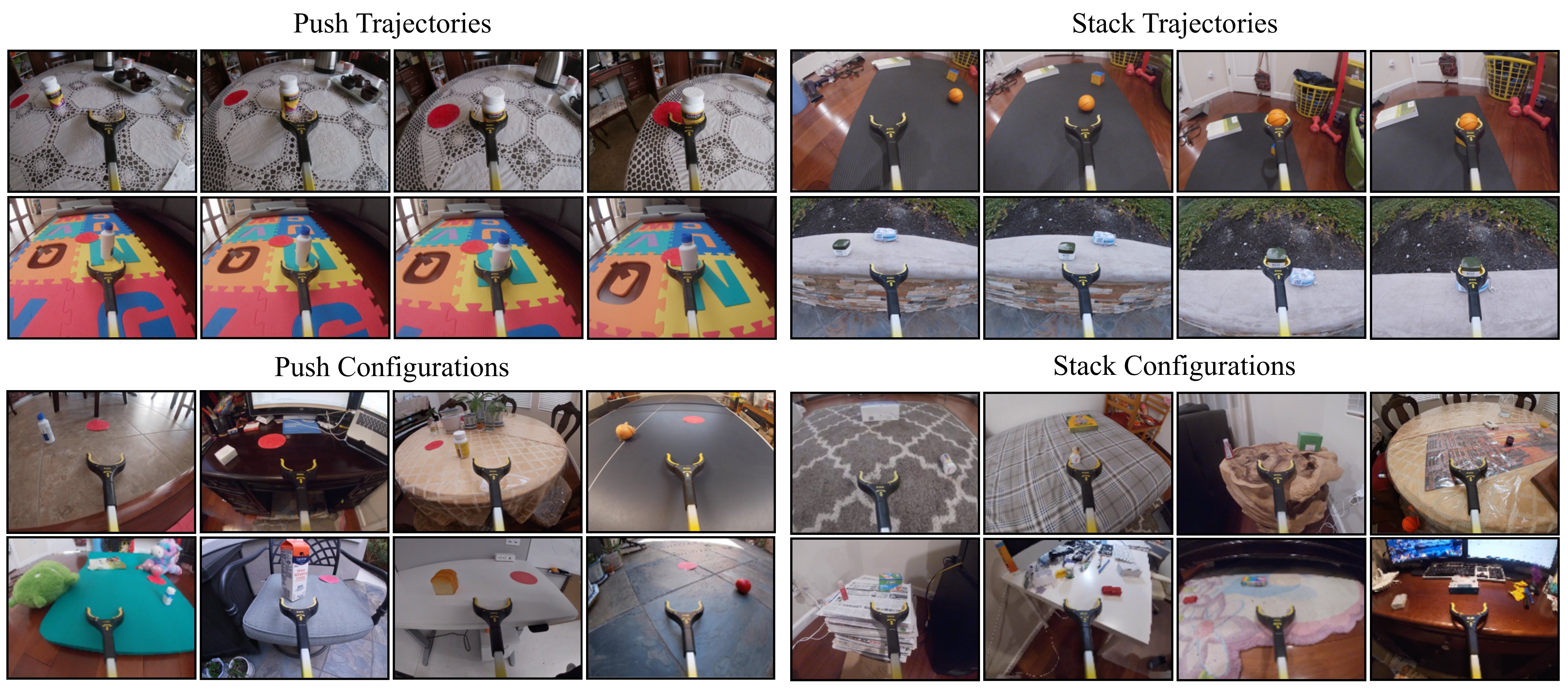}
	\caption{For both Pushing (left) and Stacking (right), we collect 1000 trajectories each with diverse objects and scenes. The top two rows depict 4 frames from single trajectories, while the bottom two rows depicts the variations in environments collected in our dataset.} 
	\label{fig:tasks}
\end{figure*}

In this section we describe our experimental evaluations using the DemoAT framework. Specifically, we aim to answer the following key questions: (a) Can DemoAT be used to solve difficult manipulation tasks? (b) How important is the scale and diversity of data for imitation learning in the wild? (c) How important is data augmentation for visual imitation? 

\subsection{Tasks}

To study the use of DemoAT, we look at two tasks, non-prehensile pushing and prehensile stacking. To evaluate our learned policy we use two metrics. First, mean squared error (BC-MSE) between predicted actions and ground truth actions on a set of held-out demonstrations that contain novel objects in novel scenes. This offline measure allows for benchmarking different learning methods. Second, we evaluate on real robot executions on previously unseen objects and measure the fraction of successful executions. This captures the ability of our learned models to generalize on real scenarios. 
% Details on specific tasks are as follows:

\textbf{Non-prehensile Pushing:} This task requires the robot to push an object to a red circle by sliding it across the table. Such contact-rich manipulation has been extensively studied and known to be challenging to solve~\cite{lynch1996nonprehensile,ruggiero2018nonprehensile}. Particularly in our case, we operate with diverse objects in diverse scenes, which makes accurately manipulating objects difficult. For robotic experiments, we evaluate robotic success rate as $\frac{\#  \textup{trajectories where object reaches goal}}{\# \textup{total trajectories}}$ on a set of 24 different objects unseen in training.

\textbf{Prehensile Stacking:} In this task, the goal is to grasp an object and stack it on top of an equally sized or larger object. We set it up such that the smaller object is always in front of the larger object to reduce ambiguity in learning (\Figref{fig:tasks}). We evaluate robotic success rate as $\frac{\#  \textup{trajectories where object is grasped and stacked}}{\# \textup{total trajectories}}$ on a set of 24 configurations unseen in training.

\begin{figure*}[t!]
	\centering
	\includegraphics[width=\textwidth]{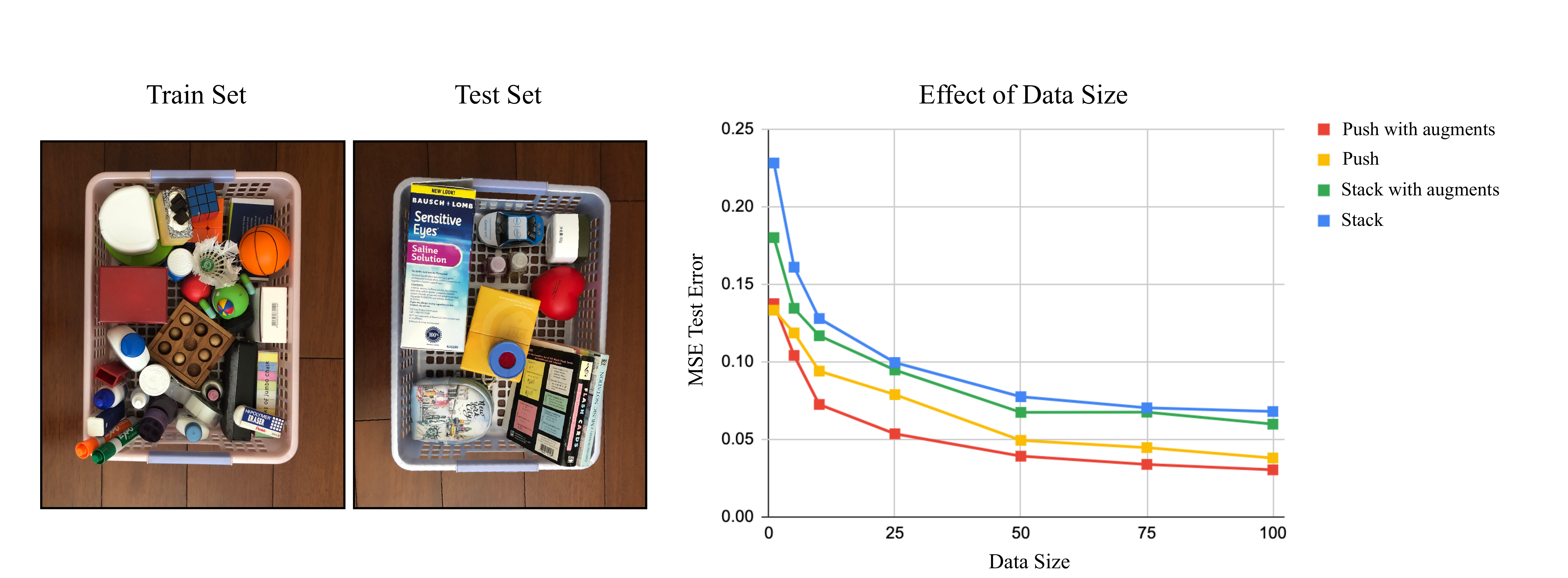}
	\caption{On the left, we show examples of objects used in training and testing for behavioral cloning evaluation. On the right, we evaluate the MSE error on held-out testing objects with and without random data augmentations. Note that as we increase the amount of data, our models improve and achieves lower error.}
	\label{fig:testrain}
\end{figure*}

\begin{table}[]
\centering
\caption{Real robot evaluation results (average success rate): Stacking is split into 2 parts for evaluation: (a) grasping the initial object and (b) stacking the object onto the second object after completing (a).}
\label{tab:real_eval}
\vspace{0.1in}
\begin{tabular}{@{}llcccc@{}}
% \toprule
\multicolumn{2}{l}{} & \begin{tabular}[c]{@{}c@{}}Naive BC\\ 100\%\end{tabular} & \begin{tabular}[c]{@{}c@{}}BC with augment\\ 100\%\end{tabular} & \begin{tabular}[c]{@{}c@{}}BC with augment \\ 50\%\end{tabular} & \begin{tabular}[c]{@{}c@{}}BC with augment\\ 10\%\end{tabular} \\ \midrule
Push & reach goal & 0.625 & \textbf{0.875} & 0.750 & 0 \\ \midrule
\multirow{2}{*}{Stack} & grasp object & 0.750 & \textbf{0.833} & 0.792 & 0 \\
 & stack object & 0.291 & \textbf{0.625} & 0.416 & 0 \\ \bottomrule
\end{tabular}
\end{table}

\subsection{Can DemoAT be used for solving difficult manipulation tasks?}
To study the utility of our DemoAT framework, we look at both measures of performance, the offline BC-MSE and the real robot success rate. Unless otherwise noted, we train our policies with 100\% of training data and using the `crop'+`jitter' augmentation for pushing and `rotate'+`jitter' for stacking (\Figref{fig:rad}). On the BC-MSE metric, we achieve an error of 0.028 on the pushing task and an error of 0.056 on the stacking task. We note that this is more than two orders of magnitude better than random actions, which has error of ~0.67 and ~0.69 on pushing and stacking respectively. This demonstrates that our policies have effectively learned to generalize to previously unseen demonstrations. Visualizations of how close predicted actions are to ground truth actions are presented in \Appref{app:predictions}.

Although our framework results in low BC-MSE error, such offline measures often do not necessarily correspond to effective online robotic performance. However, we demonstrate that our learned policies are robust enough to perform well on our robot. As seen in \Tabref{tab:real_eval}, we achieve a success rate of ~87.5\% on pushing and ~62.5\% on stacking previously unseen objects. This demonstrates that our DemoAT framework can indeed solve complex tasks in diverse domains.
% Furthermore, in the provided supplementary video on our  \href{https://sites.google.com/view/visual-imitation-made-easy}{project website}, we demonstrate how our learned closed-loop policies are robust to disturbances applied on the objects. 
%\st{should we point to project page/supplementary for demo videos?} \lp{+1 please add} \sy{added}

\subsection{How important is data for imitation learning in the wild?}
A key promise of the DemoAT setup is the ability to collect large-scale, diverse demonstrations. But how important is this diversity of data? To study this, we train policies on different fractions of training data - 1,~5,~10,~25,~50,~75,~100\% and evaluate their performance. There are two ways of creating a fractional split, either by sequentially selecting the data or by random selection. The first split will contain more data in the same environment, while the second will contain more diverse data albeit with smaller amounts for each environment. In \Figref{fig:testrain}, we use the sequential split since it better captures the process of collecting data. In \Appref{app:random_split}, we present results for random splits.

\textbf{Behavioral Cloning Evaluation:} In \Figref{fig:testrain} we illustrate the effects of changing dataset size on BC-MSE performance. In both the pushing and stacking task, we see increasing data size significantly improves performance especially in the low-data regime. We note that improvements diminish with larger data on the BC-MSE metric with just $\sim 0.9\%$ performance gain when increasing our training data from 500 to 1000 trajectories.

\begin{figure*}[t!]
	\centering
	\includegraphics[width=\textwidth]{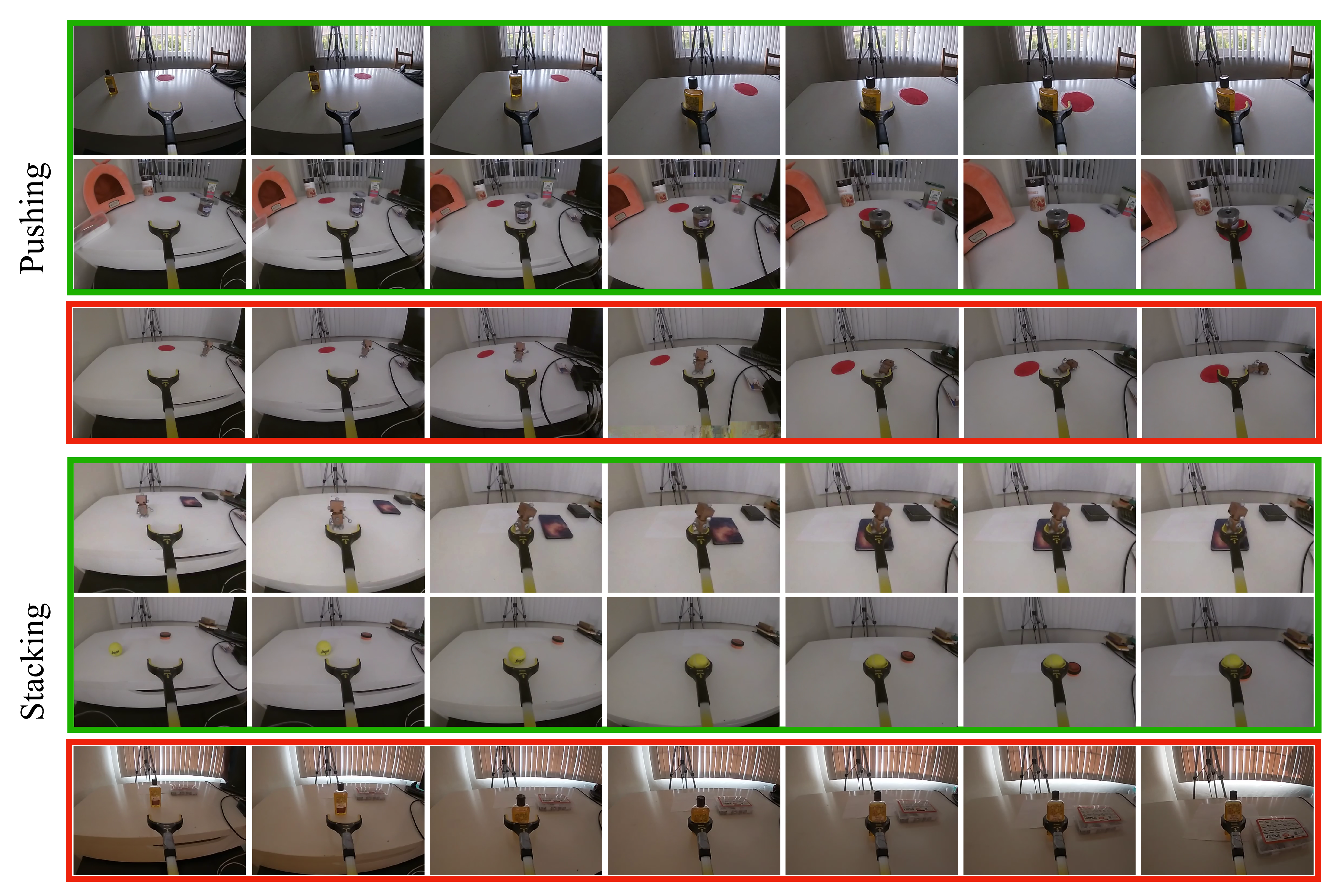}
	\caption{Here we visualize trajectories executed on the robot using our learned pushing and stacking policies trained with augmented data. Successful trajectories are highlighted in green, unsuccessful ones in red.} 
	\label{fig:fig8}
\end{figure*}

% \textbf{Behavioral Cloning Evaluation:} For both tasks, we compare performance of the best data augmentation configurations across different amounts of data: 1,5,10,25,50,75,100\%. In \Figref{} we show the error rates across both tasks.

\textbf{Real Robot Evaluation:} In \Tabref{tab:real_eval}, we report the performance of robotic execution on models trained on ~10\%, ~50\%, and ~100\% of the collected data for each task on an unseen test set of 24 different objects. In both tasks, we see that with only ~10\% of the data (100 trajectories), the robot is unable to even reach the first object. When we increase to ~50\% of the data, we see a huge improvement and the robot starts to learn to reach the objects and complete the tasks. Particularly, with just 50\% of the data, the robot can successfully reach the object ~100\% of the time in the non-prehensile pushing task. When we evaluate with all our data, we still see considerable performance improvements in completing the tasks, with ~12.5\% in pushing and ~20.9\% in stacking. This improvement is significantly higher than what we see with the BC-MSE metric. We hypothesize that since both these tasks require fine-grained manipulation, small improvements in BC-MSE results in large improvements in real-robot performance, especially when the models are already performative. 
%\lp{write more once table is ready.} 

\textbf{Data Diversity vs Size:} To further understand the effects of diversity and size, we run experiments that compare performance on the same fractional split, but different amounts of diversity in the data. Given a quota 100 demonstrations, we train on two splits of data: (A) many observations of the same objects and scenes (B) sparse observations across a diverse set of objects and scenes. We expect that dataset~(B) will be better at generalizing to unseen objects, since it sees many different scenes during training. Indeed, we find that the test error for the diverse dataset~(A) [0.081] is on average 1.4\% higher than that of dataset~(B) [0.067]. Analysis on other data splits is presented in \Appref{app:diversity_size}.
% \lp{Need to write more once the figure is ready. } \sy{there is no fig for data diversity only size? can add one in appendix?} \lp{I meant results not figure sorry}

% However, we see \lp{where?}in that during training, dataset~(B) is consistently more noisy than dataset~(A), which is likely because dataset~(A) has so few examples per environment, further highlighting the importance of both the scale of data and diversity.

\subsection{Does augmented data help?}
To improve the performance of our learned policies, we employ data augmentations in training. But, how important are these augmentations in imitation learning?

\begin{figure*}[t!]
	\centering
	\includegraphics[width=\textwidth]{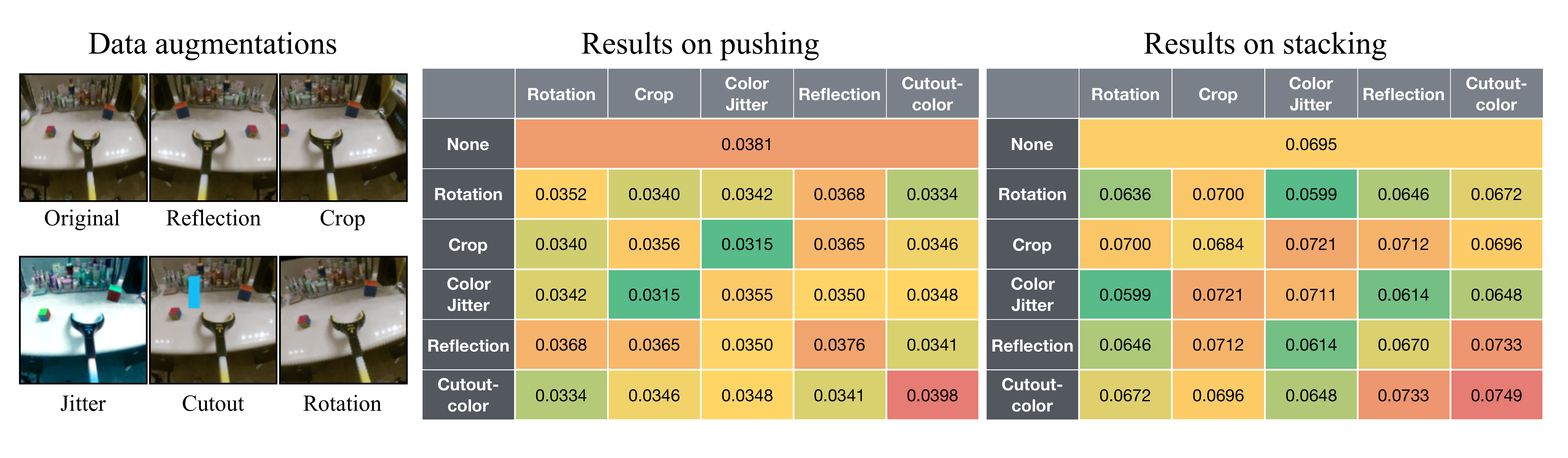}
	\caption{On the left, we show the five data-augmentations used in this work. On the middle and right, we present an analysis of MSE error (lower is better) on test-set using different combinations of data-augmentations for pushing and stacking respectively. In dark green, we highlight the best combinations.} 
	\label{fig:rad}
\end{figure*}

% We investigate whether random data augmentations improves the performance of the two tasks on unseen test sets both through a behavioral cloning error and real robot success rates. 

\textbf{Behavioral Cloning Evaluation:} For both pushing and stacking, we compare the application of different augmentations: crop, color jitter, rotate, horizontal reflection, random cutout, and all permutations of two augmentations. We find that data augmentations allow our model to generalize better to unseen objects and scenes on the BC-MSE metric. As shown in \Figref{fig:rad}, the best augmentation performs 0.7\% better than naive behavioral cloning in pushing and 0.9\% better in stacking. We note that `crop'+`jitter' is the most effective augmentation for pushing and `rotation'+`jitter' for stacking. In both tasks, random color cutout does not work as well. Since we focus on object manipulation tasks, it is likely that random color cutouts block important information such as the gripper fingers or the object, resulting in inaccurate predictions.

\textbf{Real Robot Evaluation:} Our second method of evaluation is to compare the success rates of stacking and pushing with data augmentations to naive behavioral cloning. \Tabref{tab:real_eval} shows that for both tasks we achieve significant improvements with data augmentations. We see the biggest increases in performance in the second part of each task (after the initial object has been reached): a 12.5\% improvement for reaching the goal in pushing and a 33.4\% improvement in stacking. Interestingly, using augmentations with just 50\% of training data surpasses the performance of not using augmentation with 100\% of training data on both pushing and stacking. This ability to improve performance in robotics is in line with recent research in RL~\cite{laskin2020reinforcement,drq} and computer vision~\cite{simclr}.

% \lp{write something more concrete with respect to the numbers. Mention that the same RAD pairs work well on both tasks. Some intuitive explanation of why cutout doesnt work as well compared to previous RL papers} \sy{I put this in the section above under bc eval}

\section{Conclusion}

In this paper, we present Demonstrations using Assistive Tools~(DemoAT). In contrast to traditional imitation methods that rely on domain adaptation techniques or kinesthetic demonstrations, our proposed method allows for both easy large-scale data collection and direct visual imitation learning. We learn two challenging tasks, non-prehensile pushing and prehensile stacking, and evaluate our methods via two metrics: BC-MSE and robot success rate. We have shown that using a universal reacher-grabber tool that can act as an end-effector for virtually any robot, smarter data collection methods coupled with simple behavior cloning methods and data augmentations can lead to better out of distribution performance.
We hope that this interface is a step towards more efficient robot learning, since it opens up directions for wide scale data collection and re-use. 
% The versatility of this tool makes it easy for anyone to collect data in any environment: in a lab, home, or even outdoors. 

% . This tool is capable of performing more difficult tasks, including but not limited to pouring, opening doors, and trash-picking. Furthermore, this opens up directions for wide scale data collection and re-use. The versatility of the reacher-grabber tool makes it easy for anyone to collect data in any environment: in a lab, home, or even outdoors. 

% - online data repository that pieter was talking about? \lp{we can talk about this in the conclusion, it;s hard to talk about it in experiments.}

% limitations/future work: can attach second camera below stick for more holistic view of the scene

%1) it is easier than getting kinesthetic demos

% 2) better than 3rd person imitation learning: eliminates domain gap 

% 3) this stick can attach to any robot, so it's p universal 

% 4) we randomly augment data so that in new test envs, it can generalize better

% 5) we show having diverse data is important (diff obejcts, diff scenes, etc)

% 6) we did 2 tasks: pushing (non prehensile) and stacking(prehensile) and evaluated both empirically on the real robot and looking at MSE error

\clearpage

\acknowledgments{We gratefully acknowledge the support from Berkeley Deep Drive and the Open Philanthropy Project.}

{\small
\bibliography{references}
}

\clearpage 

\appendix

\section{Details on DemoAT Demonstration Tool}
\label{app:demonstrations}
\begin{figure*}[h]
	\centering
	\includegraphics[width=\textwidth]{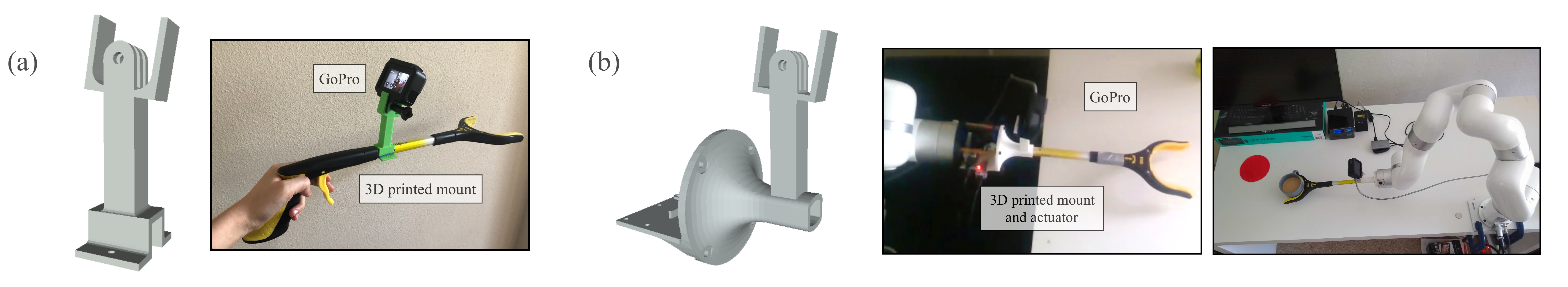}
	\caption{In part (a), we show the setup for collecting human demonstrations. This includes a 3D printed mount, a camera, and the reacher-grabber stick. Part (b) displays the corresponding setup on the robot.} 
	\label{fig:tool}
\end{figure*}

The DemoAT steup consists of the following parts:
\begin{itemize}

\item \href{https://www.amazon.com/RMS-Grabber-Reacher-Rotating-Gripper/dp/B00THEDNL8/ref=sr_1_16?dchild=1&keywords=grabber+tool&qid=1595965487&sr=8-16}{19-inch RMS Handi Grip Reacher}  
\item \href{https://gopro.com/en/us/shop/cameras/hero7-silver/CHDHC-601-master.html}{GoPro HERO7 Silver} camera 
\item 3D printed mount
\end{itemize}
Our simple setup makes it easy to start collecting demonstrations. We directly attach the angled 3D mount on the tool as shown in \Figref{fig:tool} (a) and insert the camera. This can easily be replaced by a different GoPro camera or even a phone with a modified mount. To close the fingers on the tool, the human user simply needs to press the lever. Although our tasks do not require rotating the fingers, this tool is capable of doing so.

\section{Details on DemoAT Robot End Effector}
\label{app:tool}
% should be Appendix  B
On the robot's end, we have a similar setup. We use the same reacher-grabber tool on the robot and attach it using metal studs to the robot end effector. We modified the 3D printed mount used for collecting human demonstrations with a similar one that includes an actuator to control the fingers on the tool, shown in \Figref{fig:tool} (b). 

\section{Visualizations of Predicted Actions }
\label{app:predictions}
\begin{figure*}[t!]
	\centering
	\includegraphics[width=\textwidth]{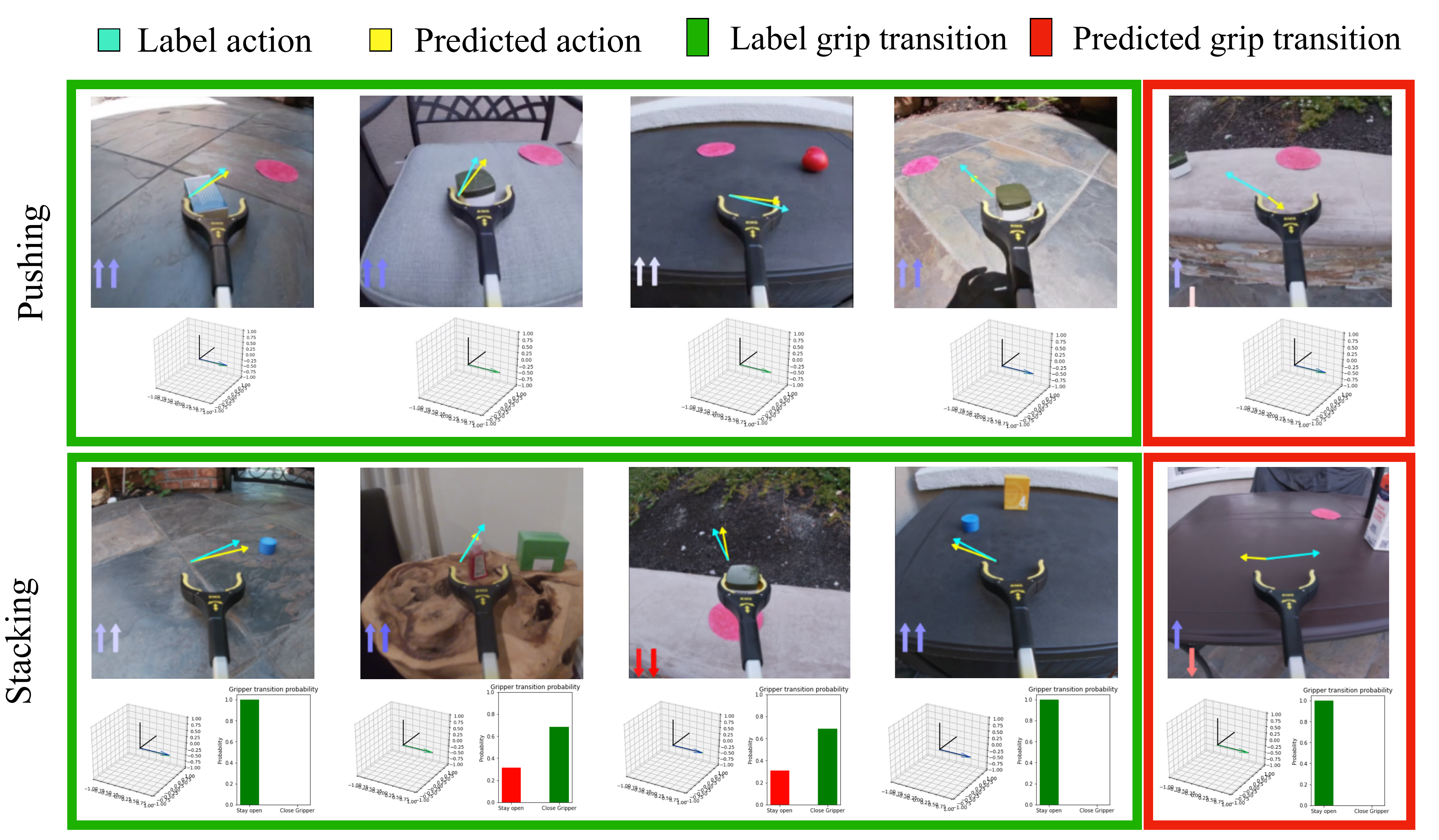}
	\caption{Examples of pushing and stacking result visualizations. The aqua arrow represents the true action across the transverse plane of the camera, while the yellow arrow represents the predicted action. The arrows in the corner are the true and predicted up-down actions. In the stacking task, the heights of the bars show the probabilities of the predicted gripper status at the next timestep, and the true gripper transition is shown in green.} 
	\label{fig:action_viz}
\end{figure*}

% should be Appendix  C 
We overlay predicted actions and COLMAP-generated labels on the images to qualitatively evaluate our results. \Figref{fig:action_viz} displays examples of test time results for both the pushing and stacking task. 

The arrows on the images show the relative translations across the transverse plane of the camera between $I_t$ and $I_{t+1}$. The aqua arrow represents the true action as output by COLMAP, and the yellow arrow represents our model's prediction. At the bottom left corner are two color-map arrows representing the up-down movement. The first arrow is the label action and the second arrow is the predicted action. The intensity of the color represents the magnitude of the action. 

The angle plot shows the predicted relative frame rotation between $I_t$ and $I_{t+1}$. The blue and green arrows represent true and predicted rotations respectively as rotation matrices multiplied by the unit vector $<1,0,0>$. We apply minimal rotation in these tasks, so these arrows are very close to $<1,0,0>$.

The bar chart in the stacking task shows the predicted probability of the status of the gripper at the next timestep. The true gripper label is green.

\section{Training Details }
\label{app:training}
% should be Appendix  D 
Let Ck denote convolutional layers with k filters and Fk denote fully connected layers of size k. 

\begin{figure*}[h]
	\centering
	\includegraphics[width=\textwidth]{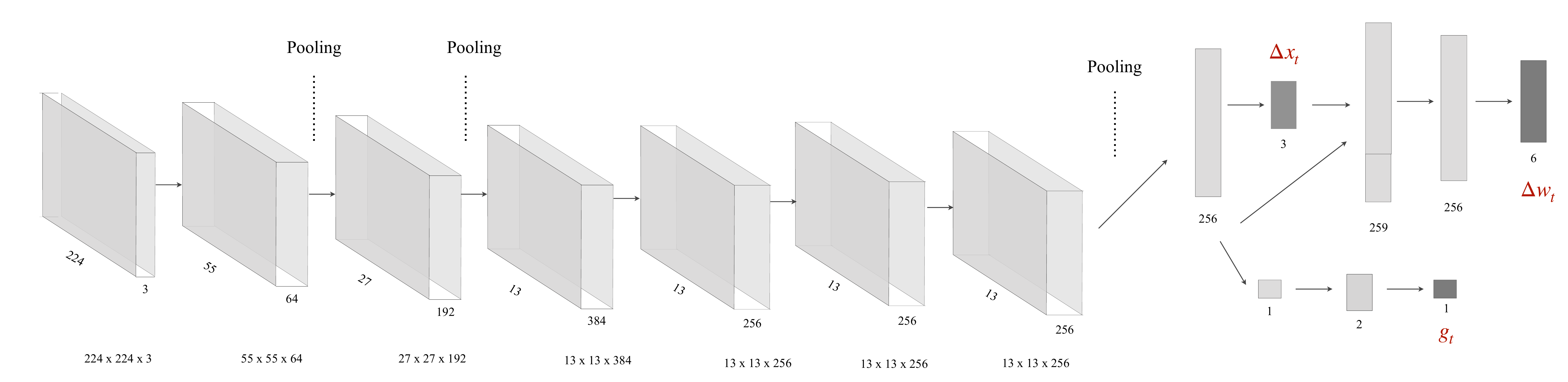}
	\caption{This is our network architecture. The input to the network is an image $I_t \in \mathbb{R}^{3 x 224 x 224}$ and it outputs $\Delta p_t = (\Delta x_t, \Delta w_t)$ and $g_t$ predictions.} 
	\label{fig:arch}
\end{figure*}

Our architecture is a set of convolutional layers followed by fully connected layers. The first part of our network takes in an image $I_t \in \mathbb{R}^{3 x 224 x 224}$ and outputs the latent representation of the observation. It consists of the first five layers of the AlexNet followed by C256 layer. We feed the latent representation into an additional net of size F512-F256 to output a relative translation vector. To get the relative rotation, we concatenate the latent representation and the translation vector and feed the result into a F256 layer before projecting to a 6D vector. This architecture is illustrated in \Figref{fig:arch}.

To train our model, we use a combination of L1, L2, and a direction loss. We care more about the direction between the prediction and ground truth actions than the magnitude, so we add the following loss to encourage this directional alignment \cite{zhang2018deep}. 
\begin{equation}
L_d =  \arccos(\frac{\Delta x_t^T \pi_\theta(\Delta x_t | o_t)}{||\Delta x_t||||\pi_\theta(\Delta x_t |o_t)||}) 
\label{eq}
\end{equation}

\section{Third-person Views of Robot Experiments}
\label{app:robot}
\begin{figure*}[h]
	\centering
	\includegraphics[width=\textwidth]{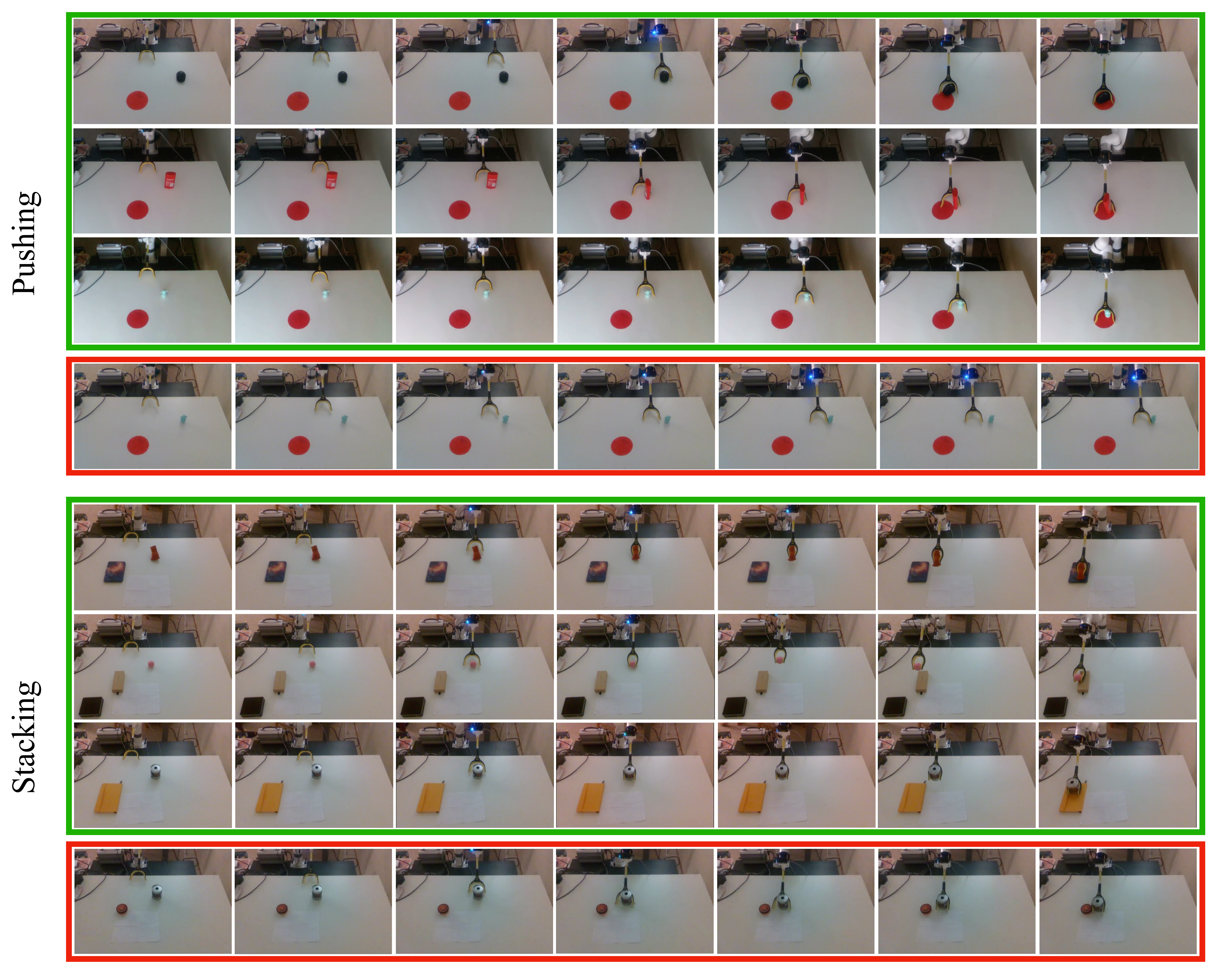}
	\caption{We visualize additional trajectories executed on the robot using our learned pushing and stacking policies from a third person point of view. Successful trajectories are highlighted in green, unsuccessful ones in red.} 
	\label{fig:more_exp}
\end{figure*}

We show additional robot trajectories for both the pushing and stacking task from a third person point of view in \Figref{fig:more_exp}. These experiments are run with the best data augmentations for each respective task. In the pushing task, the most common reason for failure is the gripper not fully wrapping around the object such that it slides out of its fingers during execution. In the stacking task, we note that common causes of failure are that the policy often grasps too late or it does not lift the object high enough to successfully stack onto the second object.

\section{Closed-loop Control with Moving Objects}
\label{app:disturbance}
\begin{figure*}[h]
	\centering
	\includegraphics[width=\textwidth]{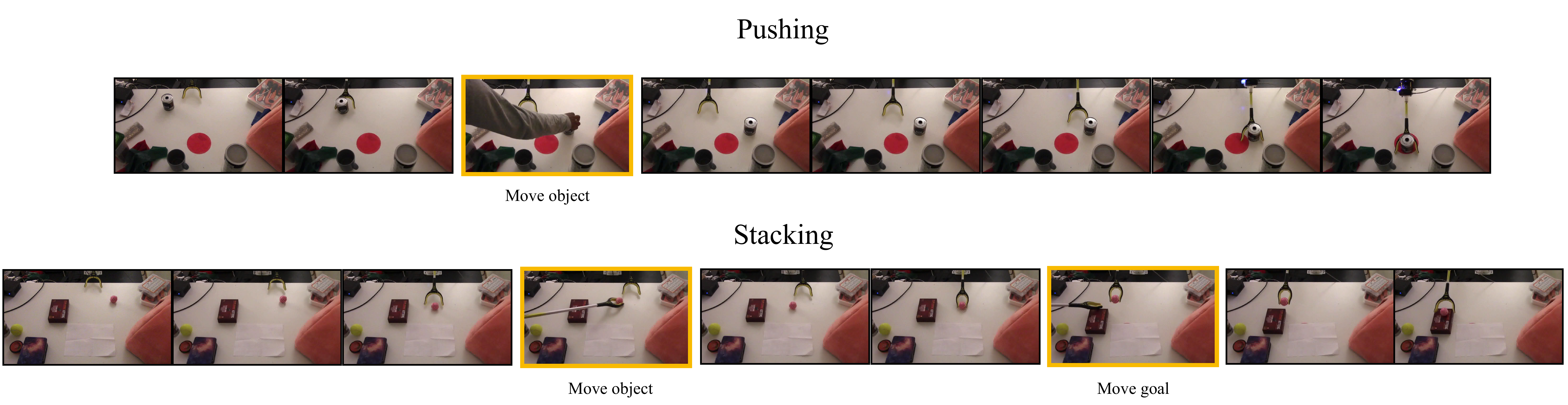}
	\caption{Here, we demonstrate how our learned closed-loop policies are robust to disturbances applied on the objects. When we slightly move the object or goal location, our policy immediately learns to adapt to the new scene. The frames where we apply a perturbation to the scene are highlighted in yellow.} 
	\label{fig:closed_loop}
\end{figure*}

We have shown that our DemoAT framework can solve complex tasks in diverse domains, and further investigate whether our learned policies are robust to disturbances. We perturb the objects and goals during online robot execution and find that our closed-loop policy is still able to successfully complete both tasks. Results are shown in \Figref{fig:closed_loop} and in the provided supplementary video on our  \href{https://sites.google.com/view/visual-imitation-made-easy}{project website}.

\section{Study of Data Augmentations on Random Data Splits}
\label{app:random_split}
\begin{figure*}[h]
	\centering
	\includegraphics[width=0.7\textwidth]{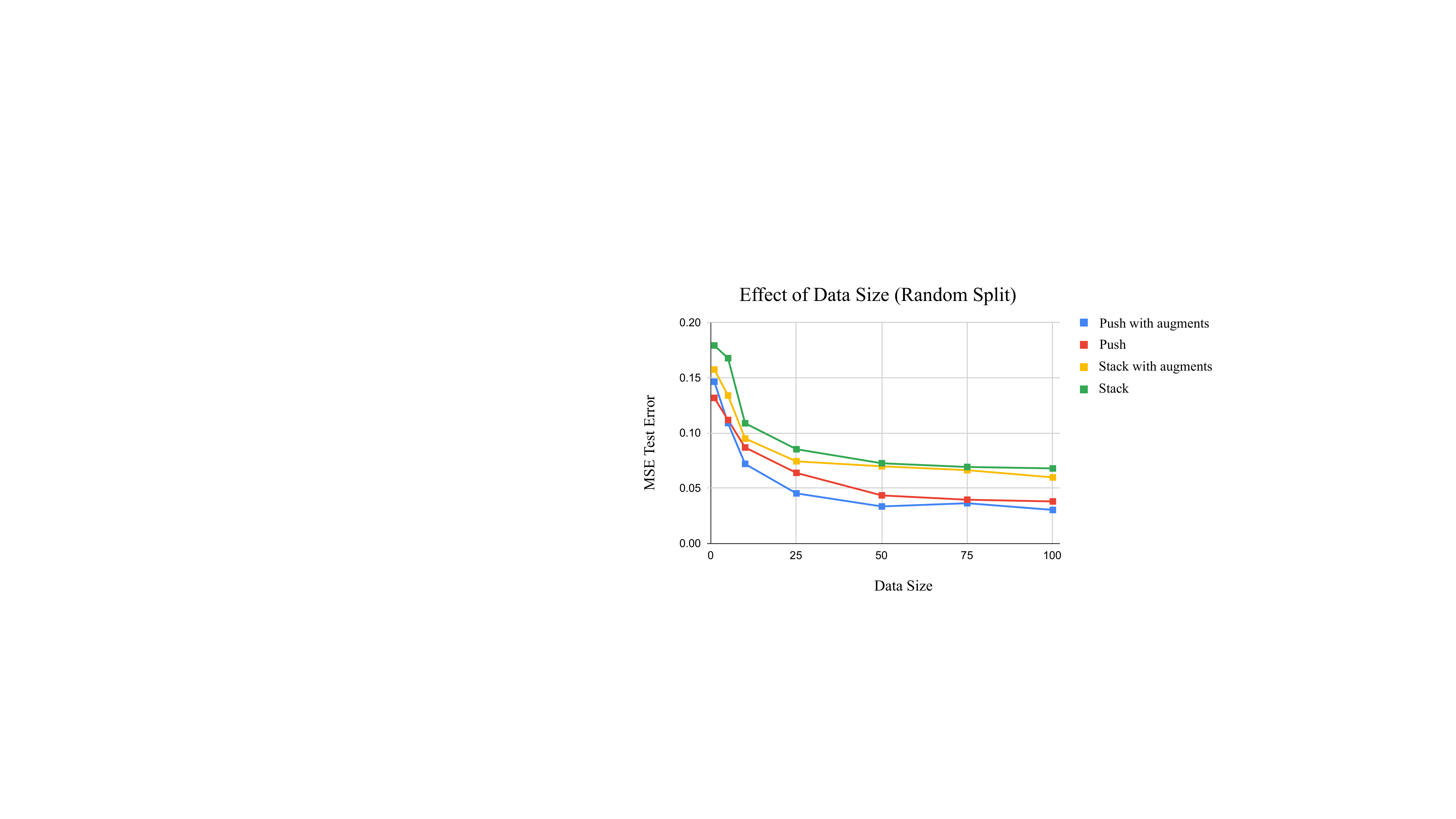}
	\caption{We show the effects of data augmentations on random diverse splits of data (instead of sequential splits). We see that regardless of how we split the data, in both tasks, data augmentations improve performance.} 
	\label{fig:random_split}
\end{figure*}

% should be Appendix  G
We show the same analysis in \Figref{fig:testrain} using random, diverse data splits instead of sequential data splits to demonstrate that data augmentations are effective and amount of data is important in both cases. Similar to the case with sequential data, we note that the performance gains diminish as we include more data in our training set. Comparing  \Figref{fig:testrain} to \Figref{fig:random_split}, we see that random splits perform better the sequential splits across every fraction of data. We analyze this difference in more detail in \Appref{app:diversity_size}.

\section{Diversity Size Analysis}
\label{app:diversity_size}
% should be Appendix  H 
We provide more detail on the results of comparing random and sequential splits. We let dataset~(A) be many observations of the same objects and scenes (sequential split) and dataset~(B) be sparse observations across a diverse set of objects and scenes (random split). We showed that in the most extreme case, when using 10\% of the data, we see an average of 1.4\% increase in performance when comparing the sequential dataset~(A) to the diverse dataset~(B). We analyze these results in more detail by running naive behavioral cloning without data augmentations for the following splits of data: 10, 25, 50, and 75\%. 

\begin{figure*}[h]
	\centering
	\includegraphics[width=\textwidth]{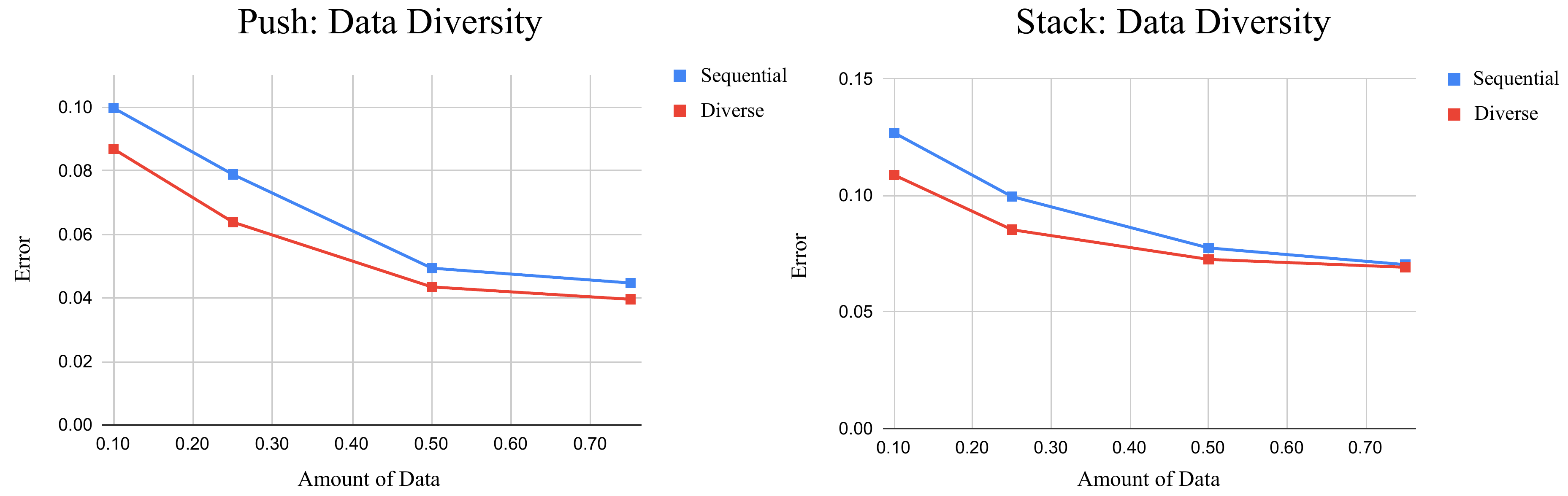}
	\caption{We show a comparison of performance between a sequential split and a random split, dataset~(A) and dataset~(B) respectively. In both tasks across all fractions of data, diverse data has much better performance.} 
	\label{fig:diversity}
\end{figure*}

In \Figref{fig:diversity}, we compare error rates for both dataset types in both pushing and stacking. The most prominent performance increase of using diverse data is when we only use 10\% of the data, and as the amount of data increases, the gap between dataset~(A) and dataset~(B) starts to decrease. Even at 75\% data, we still see a ~0.005 and a ~0.001 increase in accuracy in the pushing and stacking task, respectively. As we train on more data, it follows that the diversity of data increases and thus the difference in performance decreases.

\end{document}